\documentclass[10pt,twocolumn,letterpaper]{article}

\usepackage{iccv}
\usepackage{times}
\usepackage{epsfig}
\usepackage{graphicx}
\usepackage{amsmath}
\usepackage{amssymb}

\usepackage[breaklinks=true,bookmarks=false]{hyperref}

\iccvfinalcopy % *** Uncomment this line for the final submission

\def\iccvPaperID{****} % *** Enter the ICCV Paper ID here

\ificcvfinal\fi

\begin{document}

%%%%%%%%% TITLE
\title{S2RF: Semantically Stylized Radiance Fields}

\author{Moneish Kumar$^*$\\
\and
Neeraj Panse$^*$\\
Robotics Institute, Carnegie Mellon University\\
{\tt\small \{moneishk, npanse, dishanil\}@andrew.cmu.edu}
\and
Dishani Lahiri$^*$\\
}

\maketitle
\def\thefootnote{*}\footnotetext{Equal contribution}
\ificcvfinal\thispagestyle{empty}\fi

%%%%%%%%% ABSTRACT
\begin{abstract}
We present our method for transferring style from any arbitrary image(s) to object(s) within a 3D scene. Our primary objective is to offer more control in 3D scene stylization, facilitating the creation of customizable and stylized scene images from arbitrary viewpoints. To achieve this, we propose a novel approach that incorporates nearest neighborhood-based loss, allowing for flexible 3D scene reconstruction while effectively capturing intricate style details and ensuring multi-view consistency.
\end{abstract}

%%%%%%%%% BODY TEXT
\begin{figure*}[h]
\begin{center}
\includegraphics[width=0.70\linewidth]{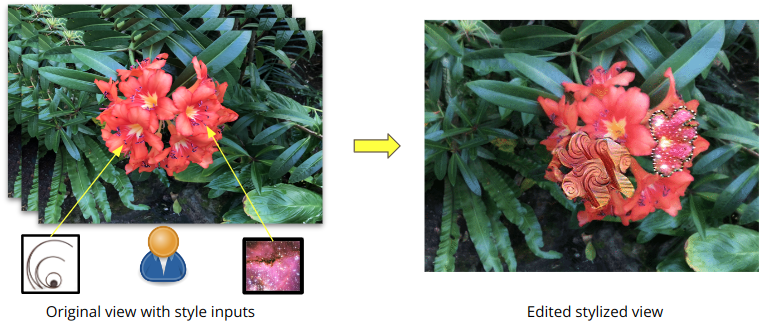}
\end{center}
   \caption{Introducing S2RF, a technique for achieving stylizable 3D reconstruction. Our method processes a set of images capturing a 3D scene and offers users the ability to style specific objects within that scene. By incorporating user-defined styles for these objects, our approach generates a stylized 3D reconstruction.}
\label{fig:main}
\end{figure*}

\section{Introduction}

For decades, recovering three-dimensional (3D) information from two-dimensional (2D) images has posed a persistent challenge in the field of computer vision. With the advent of cutting-edge differential rendering methods \cite{mildenhall2020nerf, yu2021plenoxels}, exciting new modalities have emerged, enabling the reconstruction of high-fidelity \cite{wang20234knerf, huang2023refsr, garbin2021fastnerf} and efficient \cite{chen2023mobilenerf, hu2022efficientnerf, garbin2021fastnerf} 3D scenes. 

 As advancements make 3D reconstructions more accessible, there is a growing demand for editing and manipulating these scenes. The ability to edit 3D scenes empowers creators to push the boundaries of imagination and precision. One such editing application is 3D style transfer, which aims to transfer artistic features from a single 2D image to a real-world 3D scene. Numerous remarkable works \cite{chiang2022stylizing, hollein2022stylemesh, huang2021learning, huang2022stylizednerf, nguyen2022snerf, zhang2022arf} have successfully achieved this objective. However, these methods primarily concentrate on stylizing the \textbf{whole} scene by utilizing only a \textbf{single} style image.

The primary goal of this paper is to enhance the level of control while stylizing 3D scenes. With our method, highly customizable stylized scene images can be generated from arbitrary novel viewpoints. It not only facilitates the stylization of individual object(s) but also ensures that rendered images maintain spatial consistency. Figure \ref{fig:main} provides a summary of our stylization approach. To the best of our knowledge, this is the first approach that offers a single framework for semantic and instance-level style transfer for objects in a 3D scene. 

Similar to previous methods \cite{zhang2022arf, huang2022stylizednerf, bao2023sine, zhang2023ref, hollein2022stylemesh, chiang2022stylizing},  addressing style transfer in 3D, we adopted an optimization-based approach. These methods aim to minimize (i) content loss, evaluating the difference between rendered stylized images and the original captured images, and (ii) style loss, quantifying the variance between rendered images and the style image. However, unlike these methods, we uniquely apply content and style loss exclusively to the relevant objects in the image, granting us superior control over the generated 3D scene. This approach allows for precise and targeted stylization, empowering us to achieve more tailored and refined results.

We present our results across a diverse range of 3D scenarios, showcasing how our approach serves as a stepping stone toward achieving more controllable 3D scene generation.

%-------------------------------------------------------------------------
\section{Related work}

There is a plethora of work in scene style transfer for NeRFs \cite{zhang2022arf, huang2022stylizednerf, huang2021learning, mu20223d, yin20213dstylenet, michel2022text2mesh, hollein2022stylemesh}. In the majority of style transfer pipelines, a two-staged training framework is employed. The first stage entails training a photo-realistic 3D scene, while the second stage involves fine-tuning or modifying the 3D scene representation using the style image. Some of methods represent the 3D scene in the form of meshes \cite{hollein2022stylemesh, michel2022text2mesh, yin20213dstylenet}, some in the form of point-clouds \cite{huang2021learning, mu20223d} while other using an implicit radiance field \cite{zhang2022arf, huang2022stylizednerf, mirzaei2023spin}.

These methods also differ in the way they fine-tune or modify the 3D scene representation. Some works utilize a separate hyper network \cite{chiang2022stylizing, huang2022stylizednerf} while others alter the implicit representations themselves \cite{zhang2022arf}. In the realm of 3D scene stylization, addressing spatial consistency emerges as one of the main challenges to be resolved. For example, StyleMesh \cite{hollein2022stylemesh} adopts a joint stylization approach, utilizing all input images to stylize the 3D scene and optimizing an explicit texture for accurate scene reconstruction. While Aristic Radiance field \cite{zhang2022arf} employs an exclusive color transfer to ensure view consistency. 

Controlling the style transfer and restricting it to user-specified objects is a challenging task and an active area of research. The current state-of-the-art methods perform instance-based style transfer but the quality of the renderings are not as visually pleasing and contain artifacts. An interesting work Sine \cite{bao2023sine}, requires one image from a scene edited by the user and can generate a 3D view of the scene with the edited objects. In this case, geometric priors are also used that constrain and maintain the geometric components of the objects in the scene. Albeit flexible, this method requires the user to edit one image and also the edits are semantically constrained, unlike our method where any style can latch on faithfully to an object of choice.

%-------------------------------------------------------------------------

\begin{figure*}
\label{fig:method}
\begin{center}
\includegraphics[width=0.70\linewidth]{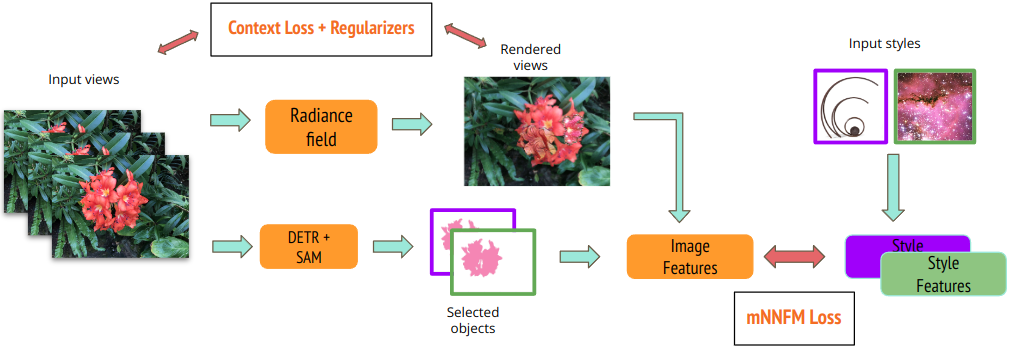}
\end{center}
   
   \caption{\textbf{Overview of our method}. We begin by reconstructing a photo-realistic radiance field and segmenting all objects from a set of scene images. Next, we apply stylization to this reconstruction by employing a masked Nearest Neighbor Feature Matching (mNNFM) style loss with the help of exemplar style images. Once the stylization process is complete, we can generate consistent free-viewpoint stylized renderings. For a more in-depth understanding of our results, we invite readers to watch the supplemental videos.}

\end{figure*}

\section{Method}

    The overview of our method is shown in Figure \ref{fig:method}. Given a set of calibrated images of a scene and a set of user-defined objects in the scene,  we aim to create a realistic and geometrically consistent image from any arbitrary viewpoint in the scene in which only the user-defined objects are styled according to reference style images. Our framework consists of three phases: generation of radiance fields, detection of objects, and stylization of radiance fields.  

\subsection{Generation of radiance field}
\label{rf_generation}

Our method uses radiance fields (RF) to represent the scene in 3D. Given a set of calibrated images of a scene, this radiance field is optimized using a rendering loss on the training rays. The method is agnostic to the way radiance fields are represented but for efficient, we use Plenoxel's \cite{yu2021plenoxels} sparse voxel grid ($\mathcal{V}$) to represent the 3D scene. Each occupied voxel stores a scalar opacity $\sigma$ and a vector of spherical harmonic coefficients for each color channel. The radiance field is defined using trilinear interpolation over sparse voxel grid.
\begin{gather}
    L(x, w) = \phi(x, \mathcal{V})
\end{gather}
Where $x$ is the queried point in the 3D space, $w$ is the queried unit directional vector, $\mathcal{V}$ is the voxel grid and the function $\phi$ is trilinear interpolation.

 It uses the differentiable volume rendering model used in NeRF \cite{mildenhall2020nerf}. The color of the ray is determined by integrating all points along the ray.
\begin{gather}
    \hat{C}(r) = \sum_{i = 1}^{N}T_{i}(1 - e^{-\sigma_{i}\delta_{i}})c_{i} \\
    T_{i} = e^{-\sum\limits_{j = 1}^{i - 1}\sigma_{j}\delta_{j}}
\end{gather}
Where $T_{i}$ represents the amount of light transmitted along the ray $r$, $\delta_{i}$ is the opacity of sample $i$, $c_{i}$ is the color of sample $i$.

Voxel grid's opacity and spherical harmonic coefficients are optimized using mean square error ($\mathcal{L}_{mse}$) over the rendered pixels along with total variation ($\mathcal{L}_{tv}$) \cite{wiles2020synsin},  beta distribution ($\mathcal{L}_{\beta}$) regularizers and sparsity prior ($\mathcal{L}_{s}$)\cite{lombardi2019neural}. The overall loss function ($\mathcal{L}_{rf}$) for the radiance field optimization is as follows:
\begin{gather}
    \mathcal{L}_{rf} = \mathcal{L}_{mse} + \lambda_{tv}\mathcal{L}_{tv} + \lambda_{\beta}\mathcal{L}_{\beta} + \lambda_{S}\mathcal{L}_{S} \label{eq:rf_loss} \\
    \mathcal{L}_{mse} = \frac{1}{|\mathcal{R}|}\sum_{r \in \mathcal{R}}||C(r) - \hat{C}(r)||^{2}_{2} \\
    \mathcal{L}_{tv} = \frac{1}{|\mathcal{V}|}\sum_{v \in \mathcal{V}}\sum_{d \in \mathcal{D}}||\Delta(v, d)||_{2}\\
    \mathcal{L}_{S} = \sum_{i}\sum_{k}log(1 + 2\sigma(r_{i}(t_{k}))^{2})\\
    \mathcal{L}_{\beta} = \sum_{r}(log(T_{FG}(r)) + log(1 - T_{FG}(r))) 
\end{gather}
Where $C(r)$ is the color of the ground truth ray, $\hat{C}(r)$ is the estimated color of the ray, $||\Delta(v, d)||_{2}$ is the squared distance between the $d$th values in the voxels. $\sigma(r_{i}(t_{k}))^{2})$ is the opacity of the sample k along the ray i. $log(T_{FG}(r))$ denotes the accumulated foreground transmittance of ray r. $\lambda_{tv}, \lambda_{\beta}$ and $\lambda_{S}$ are weights of the respective loss components.

\subsection{Detection of objects}
\label{subsection:detect_and_mask}

The second phase of our framework aims to provide a selection of objects in the scene to the user to which style can be transferred. Given a set of images of a scene, the output is a set of all object and mask pairs $\mathcal{O} = \{(o_{i}, m_{i})\}$ (where $i \in [0, N]$ and N is the number of objects) in the scene.

We use a transformer-based object detector, DEtection TRansformer (DETR) \cite{carion2020endtoend} to detect objects. Segmentation masks are obtained using the Segment Anything Model (SAM) \cite{2023SegmentA} for all images in the scene. Given an input image, DETR produces a comprehensive list of object boxes, each associated with a category tag and corresponding bounding box coordinates. SAM takes as input an image along with object bounding boxes and outputs segmentation masks corresponding to each of the input object queries. The segmentation masks generated using box prompts are much better than those generated using other prompts \cite{cheng2023sam, mazurowski2023segment} , hence we use DETR prior to SAM.   

At this point, we have a list of objects (with segmentation masks) in the scene and the corresponding style images that need to be transferred. To ensure the reliability of the detected objects, we only retain those that appear in at least 80\% of the frames throughout the scene images.

\subsection{Stylization of radiance fields}

Given a photo-realistic radiance field that is reconstructed using the method in section \ref{rf_generation} and a set of objects and masks that are obtained utilizing the approach in section \ref{subsection:detect_and_mask}, our framework finetunes the photo-realistic radiance field, in which the objects are stylized according to their respective 2D style image. We achieve this by applying the Nearest Neighbor Feature Matching (NNFM) loss \cite{zhang2022arf} to each object individually. 

The NNFM loss aims to minimize the cosine distance of each feature in the feature map of the rendered image to its nearest neighbor feature in the style images' feature map.
The rendered image from the radiance field is denoted by $I_{r}$ and the style image is denoted by $I_{s}$. The VGG feature maps extracted from both these images are $F_{r}$ and $F_{s}$ respectively. The NNFM loss is given by:

\begin{gather}
\label{eq:nnfm}
    \mathcal{L}_{NNFM} = \frac{1}{N}\sum_{i,j} \min\limits_{k,l}\delta\left(F_{r}(i,j), F_{s}(k,l)\right ) 
\end{gather}

where $F_{*}(i,j)$ denotes the feature vector at pixel location $(i,j)$ for the feature map $F_{*}$ and the function $\delta(v_1, v_2)$ computes the cosine distance between vectors $v_1$ and $v_2$.

We exclusively apply the NNFM loss (equation \ref{eq:nnfm}) to the pixels that correspond to each object separately. This selective application is achieved by employing the mask obtained in section \ref{subsection:detect_and_mask}. The mask allows to effectively confine the style transfer to the specific objects of interest. The masked NNFM  loss (mNNFM) is as follows:

\begin{gather}
    \rho = \sum\limits_{i,j} \min\limits_{k,l}m_{o}(i,j)D(F_{r}(i,j), F_{s}^o(k,l)) \\
    \mathcal{L}_{mNNFM} = \frac{1}{N}\sum_{o = 1}^{N} (\rho)
    \label{eq:nnfm_adapted}
\end{gather}

$m_o$ represents the mask specific to object $o$, while $F_{s}^o$ denotes the feature map extracted from the style image intended for transfer onto object $o$. $\rho$ represents the masked-NNFM loss over a single object and the total loss is the average over all objects in the scene.

Combining the masked NNFM loss with the loss mentioned in section \ref{rf_generation}, the overall loss that we optimize is:
\begin{gather}
    \mathcal{L} = \mathcal{L}_{rf} + \mathcal{L}_{mNNFM}
    \label{eq:nnfm_adapted}
\end{gather}
 The modified NNFM loss plays a crucial role in refining the radiance field generation process, ensuring that the applied style adheres precisely to the selected objects. This approach enhances the overall visual appeal and contextual consistency of the final output, making it more compelling and realistic.

%-------------------------------------------------------------------------

\begin{figure}
\begin{center}
\includegraphics[width=0.9\linewidth]{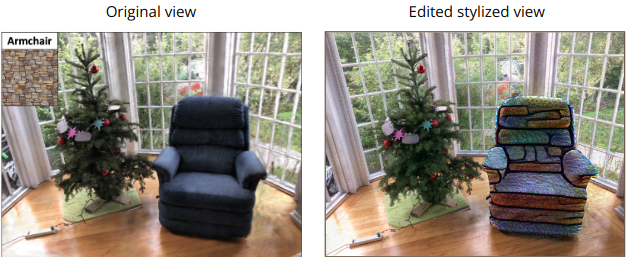}
\includegraphics[width=0.9\linewidth]{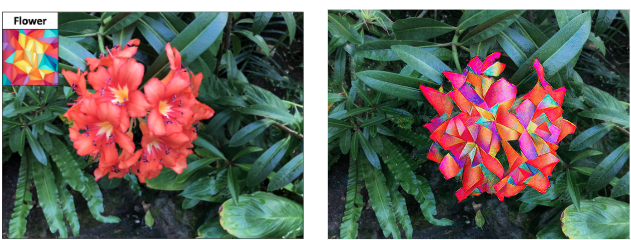}
\end{center}
   \caption{ Shows examples of stylized radiance fields in two senarios. 1)  Single object (chair) instance is stylized (Top row). 2) Multiple instances of the same object (flower) have been stylized (Bottom row). Images on the left show one of the input images for the scene along the object to be styled and style image (top-left). Images on the right show an image of the stylized image.}
\label{fig:results_1}
\end{figure}

\begin{figure}
\begin{center}
\includegraphics[width=0.9\linewidth]{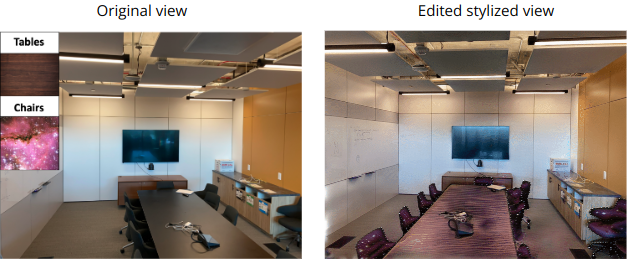}
\includegraphics[width=0.9\linewidth]{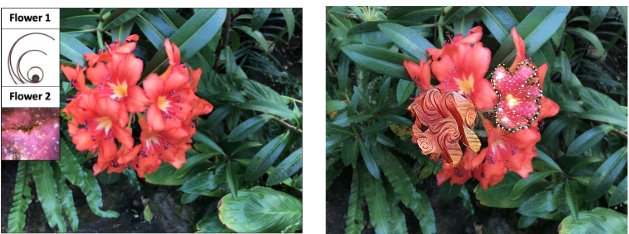}
\end{center}
   \caption{Shows examples of style transfer with 1)  Multiple instances of multiple object(s) (Chairs and Table) are stylized (Top row). 2) Multiple instances of the same object (flower) are stylized (Bottom row). Images on the left show one of the input images for the scene along the object to be styled and style image (top-left). Images on the right show an image of the stylized image.}
\label{fig:results_2}
\end{figure}

\section{Experiments}
To assess the effectiveness of our method, we conduct qualitative evaluations, showcasing results from diverse real scenes where the objects are influenced by different style images. We demonstrate how our approach successfully applies various styles to objects within real-world contexts, providing visual evidence of its versatility and performance.

\textbf{Datasets.} We conduct our experiments on multiple real-world scenes which include: \textit{Flower, Xmaschair, Room} from  \cite{mildenhall2019local}. All these scenes are front-facing captures. The style images include a diverse set of images taken from \cite{zhang2022arf}.

Our qualitative results are presented in Fig. \ref{fig:results_1},\ref{fig:results_2}. We explore four different scenarios for style transfer:

\textbf{Style transfer on a single instance of an object}: In this scenario, we apply a style image to a single object within the 3D scene. Figure \ref{fig:results_1} (Top row) shows the result of applying the style image on the chair.

\textbf{Style transfer on all instances of a single object}: In this case, we transfer a single style to all instances of a single object. Figure \ref{fig:results_1} (Bottom row) shows the application of the given style on all the flower instances in the scene.

\textbf{Style transfer on all instances of multiple objects object}: In this example, we transfer different styles to separate objects in the scene. Figure \ref{fig:results_2} (Top Row) shows the application of different styles on the table and chairs in the 3D room scene. 

\textbf{Style transfer on multiple instances of a single object}: In this case, we transfer different styles to separate instances of the same object. We apply different two different styles on two separate flowers in the same scene as shown in Figure \ref{fig:results_2} (Bottom row).

 We encourage readers to watch the supplementary videos and the appendix \ref{sec:results} to better view the results.

\section{Discussion}

We propose a novel method for reconstructing stylized radiance fields from photorealistic radiance fields. The cornerstone of our method lies in the application of masked NNFM loss, enabling a more controllable style transfer. Our method effectively achieves style transfer on both semantic and instance level, successfully applying distinct style(s) to multiple object(s) within a single scene. While this serves as a compelling proof of concept, a more comprehensive evaluation is required to fully validate our approach. Future assessments should encompass a broader range of scenes, including 360-degree environments and scenes with an increased number of objects. Additionally, it is crucial to conduct a quantitative evaluation to thoroughly assess the effectiveness of our method.

{\small
\bibliographystyle{ieee_fullname}
\bibliography{egbib}
}

\appendix

\section{Qualitative Results}
\label{sec:results}

\begin{figure}[h]
\begin{center}
\onecolumn \includegraphics[width=0.8\linewidth]{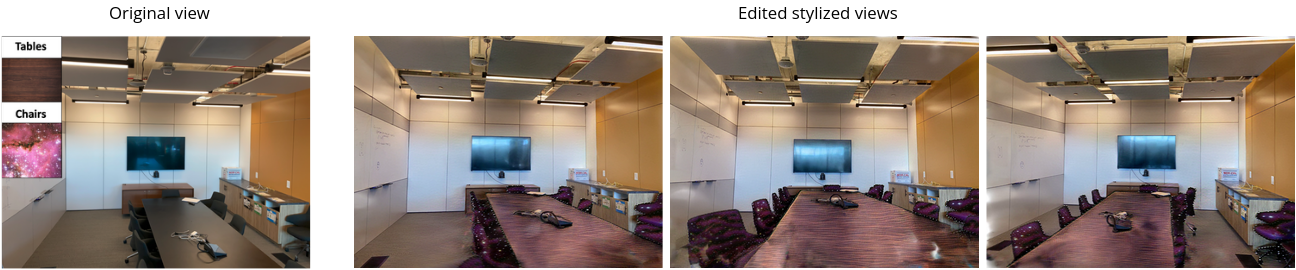}
\includegraphics[width=0.8\linewidth]{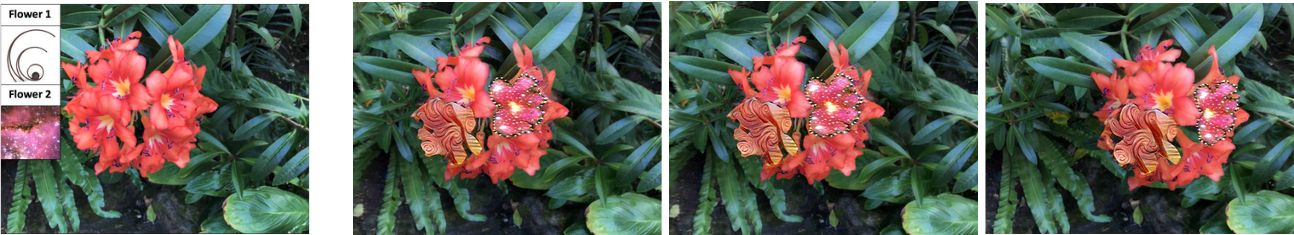}
\includegraphics[width=0.8\linewidth]{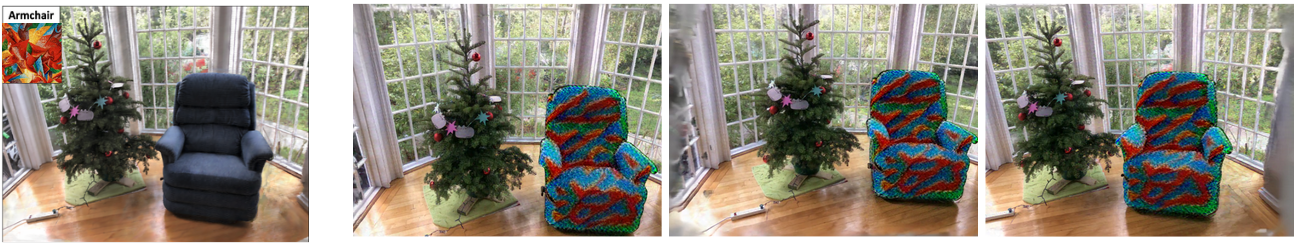}
\includegraphics[width=0.8\linewidth]{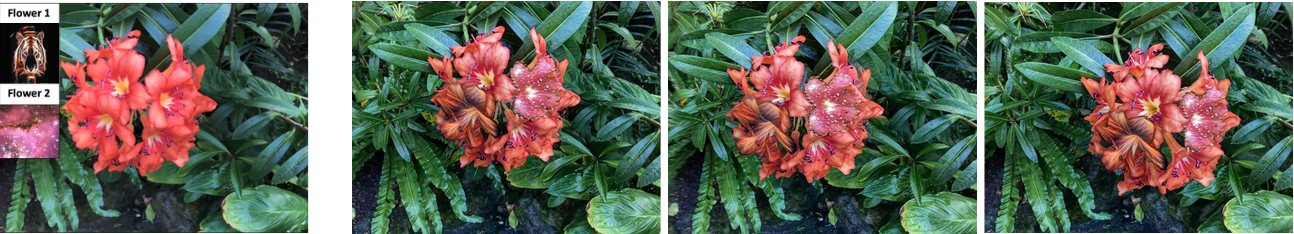}
\includegraphics[width=0.8\linewidth]{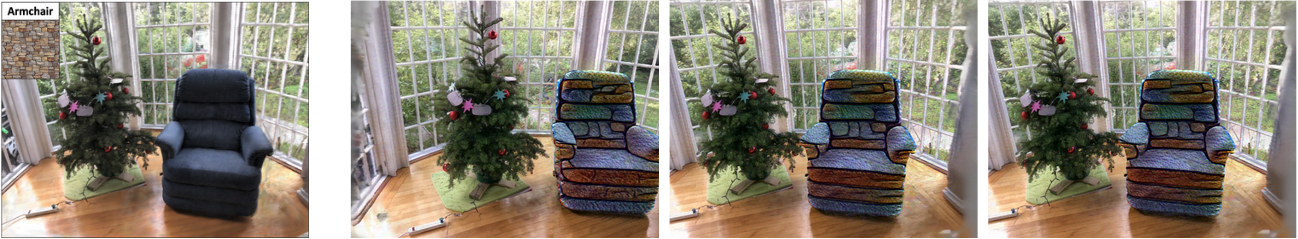}
\includegraphics[width=0.8\linewidth]{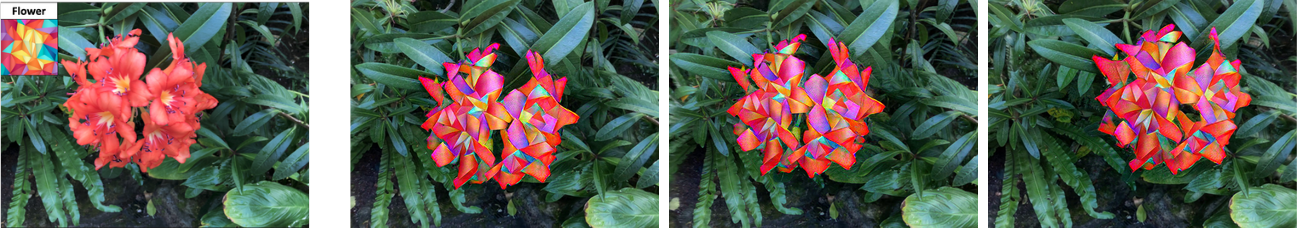}
\end{center}
   \caption{Shows qualitative results with examples of style transfers with multiple objects and style. Images in the first column show one of the input images of the scene along the object to be styled and the style image (top-left). Images on the right show three images from the stylized scene.}
\label{fig:appendix}
\end{figure}
\twocolumn
\end{document}